\documentclass[10pt,twocolumn,letterpaper]{article}

\usepackage[pagenumbers]{cvpr} 
\usepackage{makecell}
\usepackage{graphicx}
\usepackage{amsmath}

\usepackage{amssymb}
\usepackage{booktabs}
\usepackage{soul}
\usepackage{multirow}
\usepackage[accsupp]{axessibility}  
\usepackage{subcaption}

\usepackage[pagebackref,breaklinks,colorlinks]{hyperref}
\usepackage[ruled,linesnumbered]{algorithm2e}

\usepackage[capitalize]{cleveref}
\crefname{section}{Sec.}{Secs.}
\Crefname{section}{Section}{Sections}
\Crefname{table}{Table}{Tables}
\crefname{table}{Tab.}{Tabs.}

\usepackage{xcolor}
\usepackage{soul}
\setstcolor{red}

\definecolor{airforceblue}{rgb}{1.00, 0.501, 0.01}

\usepackage{enumitem}
\setlist[itemize]{leftmargin=2em}

\def\cvprPaperID{857} 
\def\confName{CVPR}
\def\confYear{2023}

\begin{document}
\title{NeRFVS: Neural Radiance Fields for Free View Synthesis\\ via Geometry Scaffolds}

\author{
Chen Yang\textsuperscript{1*}, 
Peihao Li\textsuperscript{3},
Zanwei Zhou\textsuperscript{1},
Shanxin Yuan\textsuperscript{2},
Bingbing Liu\textsuperscript{2},\\
Xiaokang Yang\textsuperscript{1},
Weichao Qiu\textsuperscript{2},
Wei Shen\textsuperscript{1\dag}
\\
\textsuperscript{1} {MoE Key Lab of Artificial Intelligence, AI Institute, Shanghai Jiao Tong University}
\\
\textsuperscript{2} {Huawei Noah’s Ark Lab}\quad
\textsuperscript{3}{Tsinghua University}\\
}
\newcommand{\nerf}{NeRF~}


\twocolumn[{
\renewcommand\twocolumn[1][]{#1}%
\maketitle
\begin{center}
    \centering
    \captionsetup{type=figure}
    \includegraphics[width=\textwidth,trim={0cm 0cm ,0cm -0.1cm},clip]{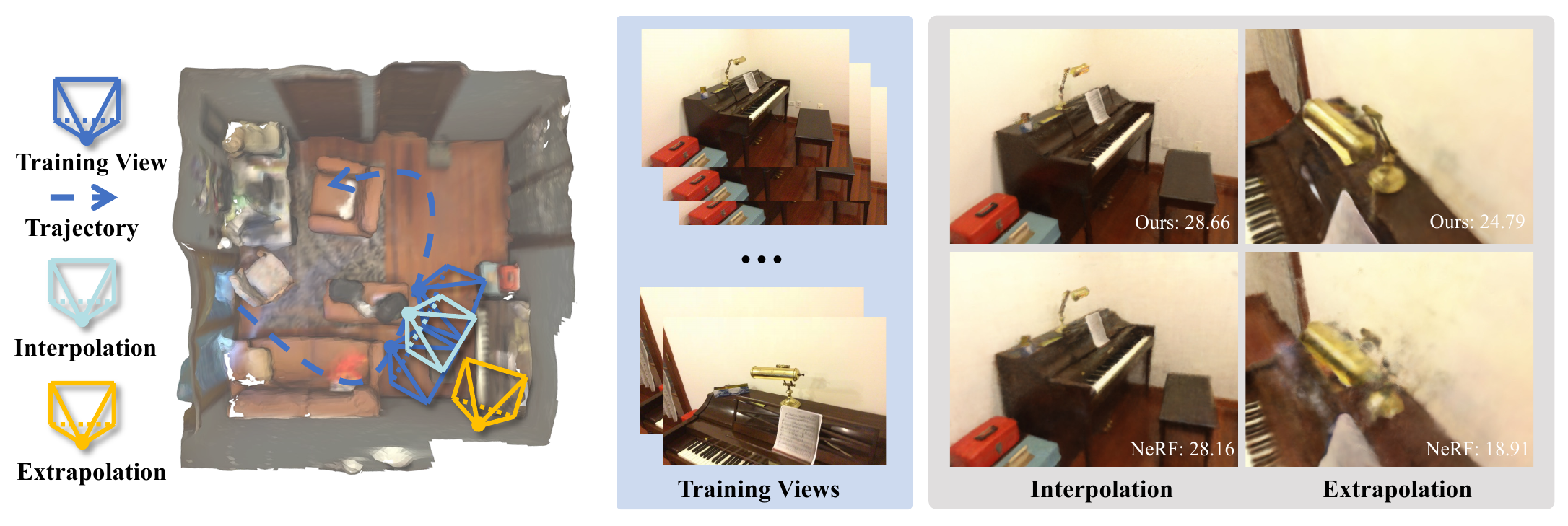}
    \captionof{figure}{\textbf{Illustration of the free view synthesis task.} 
    Free view synthesis aims at synthesizing photo-realistic images on both interpolation and extrapolation setting.
    In this paper, we propose a novel method, NeRFVS, based on the neural radiance field~(NeRF)~\cite{nerf} to achieve indoor scene free view synthesis. Our NeRFVS significantly reduces the distortions and floaters (as also evidenced by the PSNR numbers), producing high-quality images on interpolation and extrapolation setting.
    }
    \label{fig:teaser}
\end{center}
}]
\let\thefootnote\relax\footnote{$^{\textrm{*}}$ Work done during an internship at Huawei Noah's Ark Lab.}
\let\thefootnote\relax\footnote{$^{\textrm{\dag}}$Corresponding Author.}

\begin{abstract}
We present NeRFVS, a novel neural radiance fields (NeRF) based method to enable free navigation in a room. NeRF achieves impressive performance in rendering images for novel views similar to the input views while suffering for novel views that are significantly different from the training views.
To address this issue, we utilize the holistic priors, including pseudo depth maps and view coverage information, from neural reconstruction to guide the learning of implicit neural representations of 3D indoor scenes.
Concretely, an off-the-shelf neural reconstruction method is leveraged to generate a geometry scaffold. 
Then, two loss functions based on the holistic priors are proposed to improve the learning of NeRF: 1) A robust depth loss that can tolerate the error of the pseudo depth map to guide the geometry learning of NeRF; 2) A variance loss to regularize the variance of implicit neural representations to reduce the geometry and color ambiguity in the learning procedure. These two loss functions are modulated during NeRF optimization according to the view coverage information to reduce the negative influence brought by the view coverage imbalance.
Extensive results demonstrate that our NeRFVS outperforms state-of-the-art view synthesis methods quantitatively and qualitatively on indoor scenes, achieving high-fidelity free navigation results.

\end{abstract}
\section{Introduction}

\begin{figure}[t]
\centering
\includegraphics[width=\linewidth]{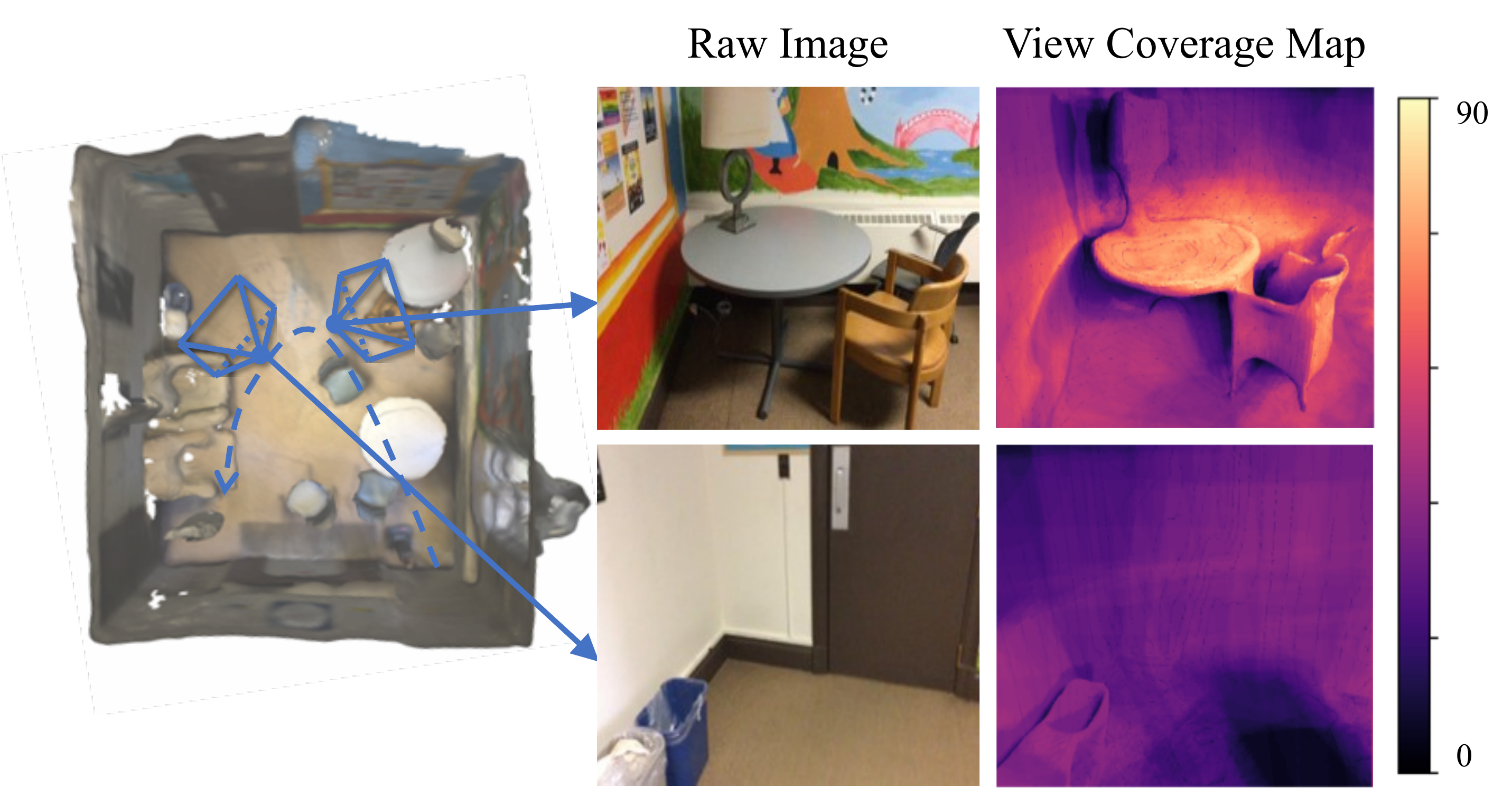}
\caption{\textbf{Imbalanced view coverage for an indoor scene.} 
The table is captured far more times than the floor and wall. 
}
\vspace{-\baselineskip}
\label{fig:view imbalance}
\end{figure}

Reconstructing an indoor scene from a collection of images and enabling users to navigate inside it freely is a core component for many downstream applications. 
It is the most challenging novel-view-synthesis (NVS) task, since it requires high fidelity synthesis from \textit{any view}, including not only views similar to the training views (interpolation), but also views that are significantly different from input views (extrapolation), as shown in Fig.~\ref{fig:teaser}. 
To clarify its difference to other NVS tasks, we term it as free-view-synthesis (FVS).
The difficulties of FVS lie in not only the common obstacles in scene reconstruction, including low-texture areas, complex scene geometry, and illumination change, but also view imbalance, \emph{e.g.}, casual photos usually cover the scene unevenly, with hundreds of frames for one table and a few for the floor and wall, as shown in Fig.~\ref{fig:view imbalance}.

Recently, \nerf has emerged as a promising technique for 3D reconstruction and novel view synthesis. 
Although \nerf can achieve impressive interpolation performance, its extrapolation ability is relatively poor~\cite{rapnerf}, especially for low-texture and few-shot regions. 
In contrast, some neural reconstruction methods can recover the holistic scene geometry successfully with various priors~\cite{manhattansdf, monosdf, neuris, neuralrecon}, while the synthesized images from these methods contain plenty of artifacts and are over-smoothed.
Inspired by the phenomena, we demonstrate that equipping the NeRF with the scene priors of the geometry captured from neural reconstruction is a potential solution for indoor FVS.

Extending \nerf to enable FVS with geometry from neural reconstruction methods is a non-trivial task with two main challenges. 
1) \textit{Depth error}. The reconstructed geometry might contain some failures, including holes, depth shifting, and floaters. The optimization of \nerf relies on the multi-view color consistency, while these failures may conflict with the multi-view color consistency, resulting in artifacts.
2) \textit{Distribution ambiguity}. 
The depth from NeRF is a weighted sum of sampling distance. 
Merely supervising the depth expectation leads to arbitrary radiance distribution, especially in low-texture and few-shot regions. This ambiguous distribution leads to floaters and blur among rendered images, as shown in Fig.~\ref{fig:var_demo}.

In this paper, we propose a novel method which exploits the holistic priors, including pseudo depth maps and view coverage information, outputted from a geometry scaffold to guide NeRF optimization, significantly improving quality on low-texture and few-shot regions.
Specifically, to address the \textit{depth error}, we propose a robust depth loss that can tolerate the error from the pseudo depth maps, reducing the negative impact of inaccurate geometry. 
As for the \textit{distribution ambiguity}, it mainly happens in the low-texture and rarely observed areas, \textit{e.g.} ceilings. We propose a variance loss to regularize the variance of the density and color distribution to decrease the ambiguity of these areas. The weights of these two losses are further adjusted according to the view coverage sufficiency to reduce the negative influence brought by the view imbalance.
With the geometry priors and variance regularization, our method can significantly reduce the floaters and distortions among low-texture and few-shot regions, achieving high-fidelity extrapolation performance. 

Experiments on synthetic and real-world datasets demonstrate that our method performs high-fidelity extrapolation by removing the distortions and floaters, significantly outperforming other view synthesis methods. 
Considering the rendering quality and 3D consistency among interpolation and extrapolation, our NeRFVS achieves new state-of-the-art performance on indoor scene FVS.

In conclusion, our contribution can be summarized as follows:
\begin{itemize}
\setlength\itemsep{-.3em}
\item A novel approach enabling neural radiance fields to perform free view synthesis on real-world scenes at room scale.
\item A robust depth loss to address the inaccuracy of the neural-reconstructed geometry.
\item A flexible variance loss with view coverage based adjustment to improve the rendering quality among low-texture and few-shot regions.
\end{itemize}

\cvprsection{Related Work}










\begin{figure*}[t]
\centering
    \includegraphics[width=\linewidth]{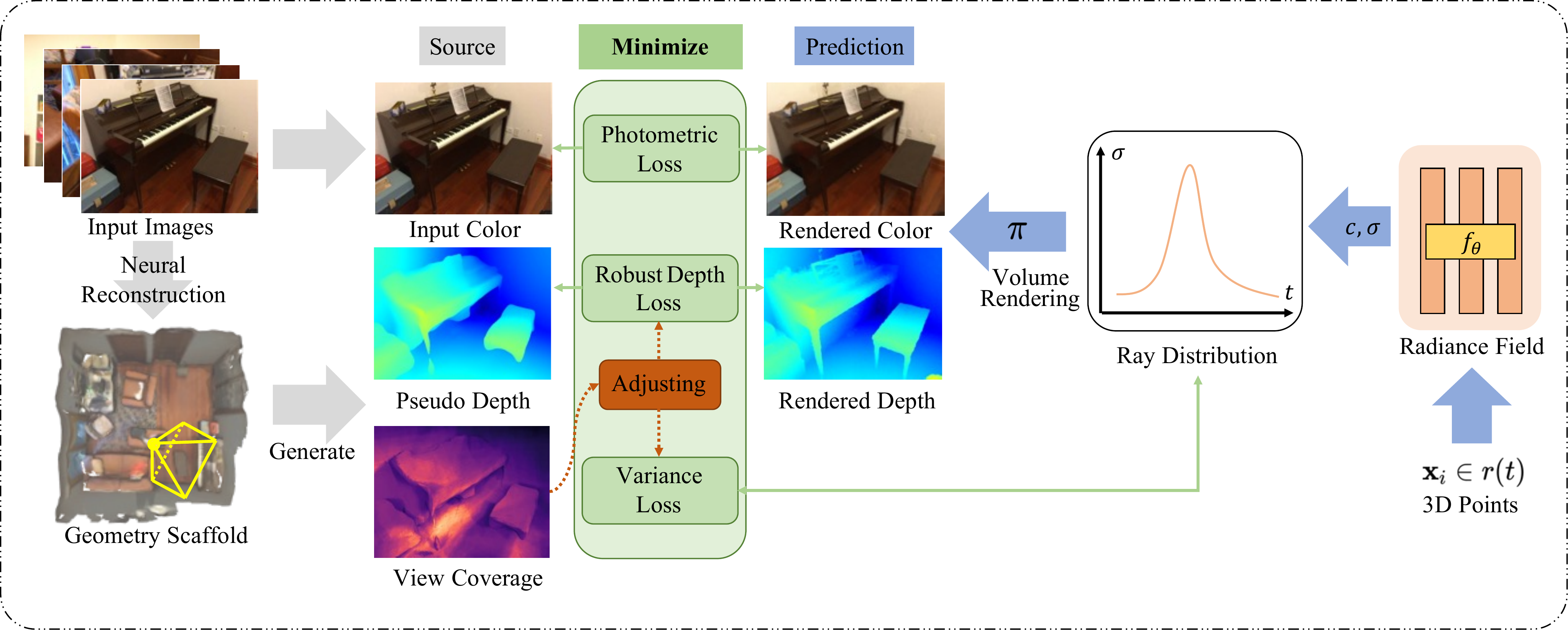}
    \caption{\textbf{Overview of NeRFVS.} 
    From left to right, we first generate a geometry scaffold via an off-the-shelf neural reconstruction method, then we get the pseudo depth and view coverage map from the scaffold.
    From right to left, we optimize a NeRF with regularization from color, depth, and variance distribution along with adjustments based on view coverage.
    }
    \vspace{-\baselineskip}
\label{fig:pipeline}
\end{figure*}

\noindent \textbf{Indoor Scene Rendering}
A classical type of method is image-based rendering (IBR)~\cite{ulr}, which fuses multi-view images to output a target view image. FVS~\cite{fvs} and SVS~\cite{svs} created a coarse geometric scaffold via Structure from Motion (SfM)~\cite{colmap} and applied an encoder-decoder structure to extract features and map them to target views for synthesis. However, these methods fail when few views are provided. SIBRNet~\cite{sun2022learning} proposed to recover geometry from sparse depth by depth completion. Some methods focus on indoor lighting. Given a 3D mesh obtained by multi-view stereo (MVS) reconstruction, Philip~et al.~\cite{philip2021free} employed an implicit representation of scene materials and illumination to enable relighting and free-viewpoint navigation. PhotoScene~\cite{yeh2022photoscene} inferred spatially-varying procedural materials and scene illumination from source images. DSRNet~\cite{xu2021scalable} extracted a global mesh from input images and used a two-layer geometric representation to encode RGB-D and reflection information. Recently, some methods combined volume rendering into their pipeline. IBRNet~\cite{wang2021ibrnet} fed features from source images into an MLP backbone and produced point-wise color and density values for volume rendering. Scalable neural scene~\cite{wu2022scalable} allocated tiles on a global mesh proxy. Each tile encoded surface and reflection representations, which were summed via volume rendering to produce the final rendering results. Compared with previous methods, our method achieves indoor scene free navigation without ground truth mesh nor RGB-D sensors.

\noindent \textbf{NeRF with Depth and Regularization}

NeRF~\cite{nerf} generates accurate geometry under dense multi-view supervision, but struggles with sparse views and unbounded regions, leading to inaccurate depth values and ``floaters'' in renderings. Previous methods have proposed regularization strategies to address this issue, such as explicit density constraints in InfoNeRF~\cite{kim2022infonerf} and weight consolidation in Mip-NeRF 360~\cite{barron2022mip}, or depth priors in RegNeRF~\cite{niemeyer2022regnerf}, SinNeRF~\cite{xu2022sinnerf}, DS-NeRF~\cite{deng2022depth}, and Dense Depth Priors~\cite{roessle2022dense}. Our method constrains both depth expectation and density distribution for more consistent visual effects.

\cvprsection{Preliminaries}
We make a quick review of NeRF's rendering pipeline. A NeRF, optimized by multi-view consistency, represents a 3D scene as the volume density and the directional emitted radiance for any point in space. Given a camera ray parameterized as $\mathbf{r}(t) = \mathbf{o} + t\mathbf{d}$ that passes through the 3D scene, the color and depth are rendered as follows: 
\begin{align}
    \hat{\mathbf{C}}(\mathbf{r})&=\int_{t_{n}}^{t_{f}} w(\mathbf{r}(t)) \mathbf{c}(\mathbf{r}(t),\mathbf{d}) dt ,
    \label{eq:volume_render}\\
    \hat{D}(\mathbf{r})&=\int_{t_{n}}^{t_{f}} w(\mathbf{r}(t)) t dt ,
    \label{eq:depth_render}
\end{align}
where $w(\mathbf{r}(t))=T(t) \sigma(\mathbf{r}(t))$, $t_{n}$ and $t_{f}$ represent the near and far bounds, respectively. The weights $w(t)$ are computed by the accumulation of transmittance  $T(t) = \exp{(-\int_{t_{n}}^{t}\sigma(\mathbf{r}(s))ds)}$, which stands for the accumulated transmittance from $t_{n}$ to $t$. $\sigma$ and $\mathbf{c}$ represent the volume density and emitted color respectively. 
We refer to the loss function used by NeRF as the photometric loss $\mathcal{L}_{\text{color}}$:
\begin{align}\label{eq:photometric_loss}
\mathcal{L}_{\text{color}}(\mathbf{r}) & = \|\hat{\mathbf{C}}(\mathbf{r})-\mathbf{C}(\mathbf{r})\|_{2}^{2},
\end{align}
where the $\mathbf{C}(\mathbf{r})$ represents the color of the ray $\mathbf{r}$. 

Recently, instant-NGP~\cite{ngp} proposed to represent the whole 3D space with multi-resolution grids stored in a hash table.
Volume rendering is also applied to sum discrete values in each ray and generate the rendering results.
Thanks to this efficient structure, instant-NGP accelerates the training and inference stage of \nerf by a large margin without obvious performance degradation.
Our method can be easily integrated into instant-NGP and boost its extrapolation performance on indoor scene rendering as well.

\cvprsection{Method}

Our method facilitates room-scale free view synthesis from a human captured collection of RGB images $\left\{I_{i}\right\}_{i=0}^{N-1}$, 
along with cameras' intrinsic parameters $\mathbf{K}_{i} \in \mathbb{R}^{3 \times 3}$ and poses $\mathbf{p}_{i} \in \mathbb{R}^{6}$. 
We first generate a geometry scaffold from off-the-shelf geometry reconstruction methods \cite{manhattansdf} using the images and camera parameters (Sec.~\ref{ssec: Geometry Scaffold}).
Then the scene priors from the scaffold are integrated into the optimization of a neural radiance field with a robust depth loss to alleviate the influence of scaffold error (Sec.~\ref{ssec: Robust Depth Loss}). 
To decrease the ambiguity of low-texture areas, we regularize the variance of predicted density distribution and color values along per ray (Sec.~\ref{ssec: Mean-Var Regularization}).
Finally, we propose a training strategy based on the view coverage of each input view to adjust the intensity of the depth and variance loss accordingly (Sec.~\ref{ssec: View Count Adjustment}).
Fig.~\ref{fig:pipeline} shows an overview of our method.

\subsection{Geometry Scaffold and View Coverage}
\label{ssec: Geometry Scaffold}
Our method applies a 3D geometric scaffold to guiding the optimization of Nerf. To construct this scaffold, we use the geometry from neural geometry reconstruction methods~\cite{imap, Nice-slam, manhattansdf} as they can reconstruct a complete and smooth global mesh containing holistic priors.
Specifically, we apply the geometry produced by~\cite{manhattansdf} which incorporates Manhattan world assumptions on the structure of the scene into the optimization process. 
\cite{manhattansdf} achieves smooth and coherent geometry reconstruction with semantic information, especially in planar regions.

We can render depth maps and compute the view coverage information with the geometry scaffold.
We convert the scene priors from the scaffold into the depth priors by rendering distance maps $\{\mathbf{D}_i\}_{i=0}^{N-1}, \mathbf{D}_i \in \mathbb{R}_{+} ^ {H \times W \times 1}$. Every pixel's value in the distance map represents the distance, rather than depth, to the camera.
Besides, we generate view coverage maps $\{\mathbf{V}_i\}_{i=0}^{N-1}, \mathbf{V}_i \in [1, N] ^ {H \times W \times 1}$ by ray casting and shadow mapping. Concretely, we cast rays among test views and get the corresponding positions. Then we re-project these points into training views and use the shadow mapping technique to consider occlusion. In this manner, every pixel's value of $\mathbf{V}_i$ represents the observed times of this pixel among the training set.

\subsection{Robust Depth Loss}
\label{ssec: Robust Depth Loss}

\begin{figure}[t]
\centering
\includegraphics[width=\linewidth]{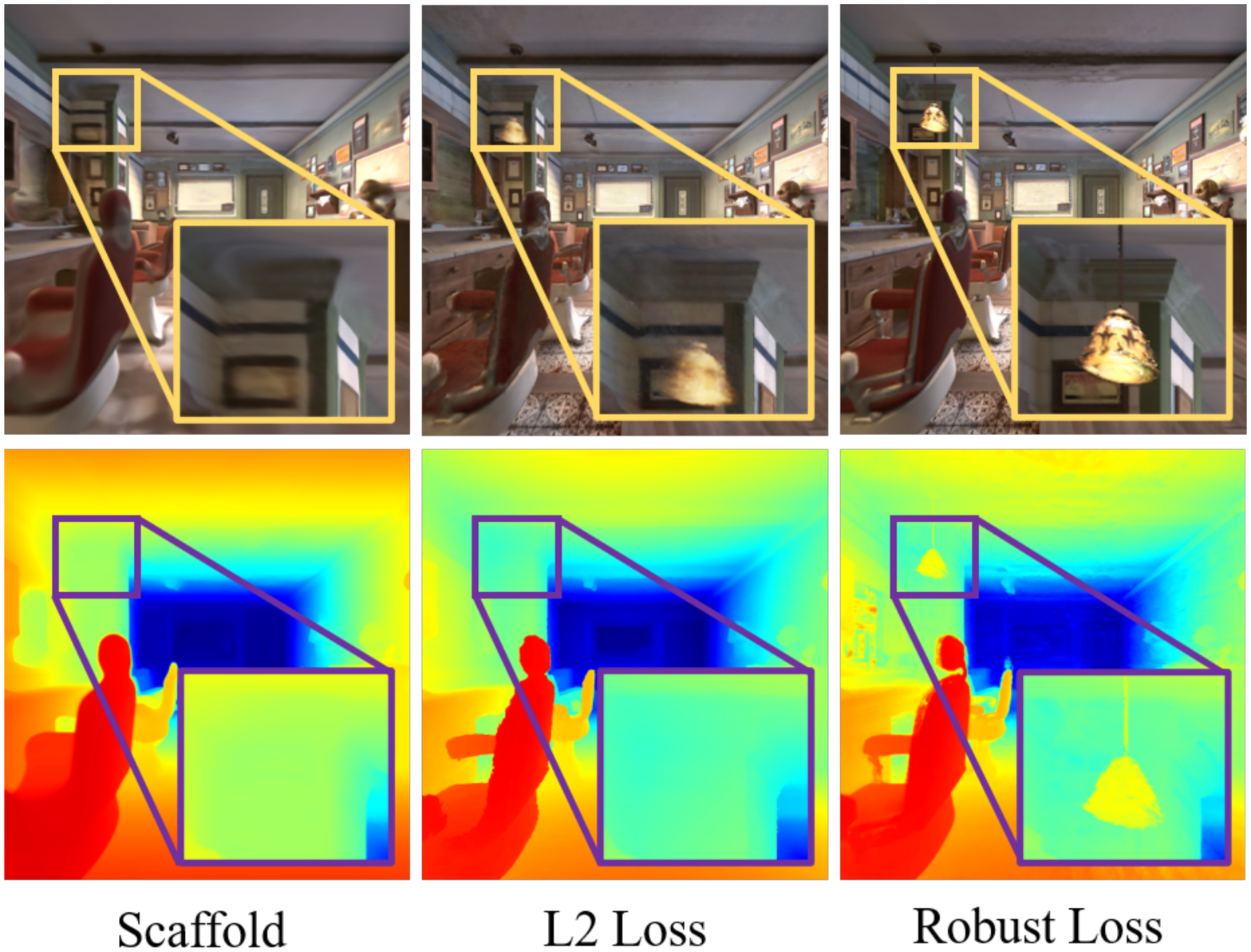}
\vspace{-2\baselineskip}
\caption{\textbf{Effectiveness of our robust depth loss}. Using the proposed robust loss, our method can recover more accurate geometry even when the depth map is inaccurate, \textit{e.g.}, it helps to recover the ceiling lamp from the wrong geometry scaffold.
}
\label{fig:robust_loss}
\vspace{-\baselineskip}
\end{figure}

In contrast to sparse and confident depth priors, we optimize \nerf with dense while inaccurate depth maps. 
If we directly adopt the optimization strategy from previous methods combining NeRF and geometry, the geometry error would degrade the extrapolation performance, as shown in Fig.~\ref{fig:robust_loss}.
Inspired by~\cite{fastrcnn, huberloss}, we propose a robust depth loss to address this issue.
Given a ray $\mathbf{r}(t) = \mathbf{o} + t\mathbf{d}$  and its distance value $D(\mathbf{r})$ from the geometry scaffold, our robust loss is defined as:
\begin{align}\label{eq:rubust depth loss}
\mathcal{L}_{\text {robust}}(\mathbf{r}) & = \begin{cases}
0.5\Delta{D}(\mathbf{r})^{2}, &\text{if } \Delta{D}(\mathbf{r}) < \beta \\
\beta ^ {2} (0.5 + \log(\frac{\Delta{D}(\mathbf{r})}{\beta})), &\text {otherwise }
\end{cases}
\end{align}
where $\Delta{D}(\mathbf{r})=\left|\hat{D}(\mathbf{r})-D(\mathbf{r})\right|$ represents the absolute depth difference, $\beta$ is a constant set to 0.1 throughout the experiments, and $\hat{D}(\mathbf{r})$ is the depth expectation along the camera axis of $\mathbf{r}(t)$.

The region where the absolute depth difference is larger than $\beta$ makes the robust loss less sensitive to outliers than L2, with corresponding gradient gradually decreasing to zero.
The $\mathcal{L}_{\text {robust}}$ encourages \nerf to terminate rays around the pseudo depth. At the same time, \nerf can retain some freedom for inaccurate geometry since the gradient gets negligible when the absolute depth difference is approaching zero or very large.

\subsection{Variance Regularization}
\label{ssec: Mean-Var Regularization}
\begin{figure}[t]
\centering
\includegraphics[width=\linewidth]{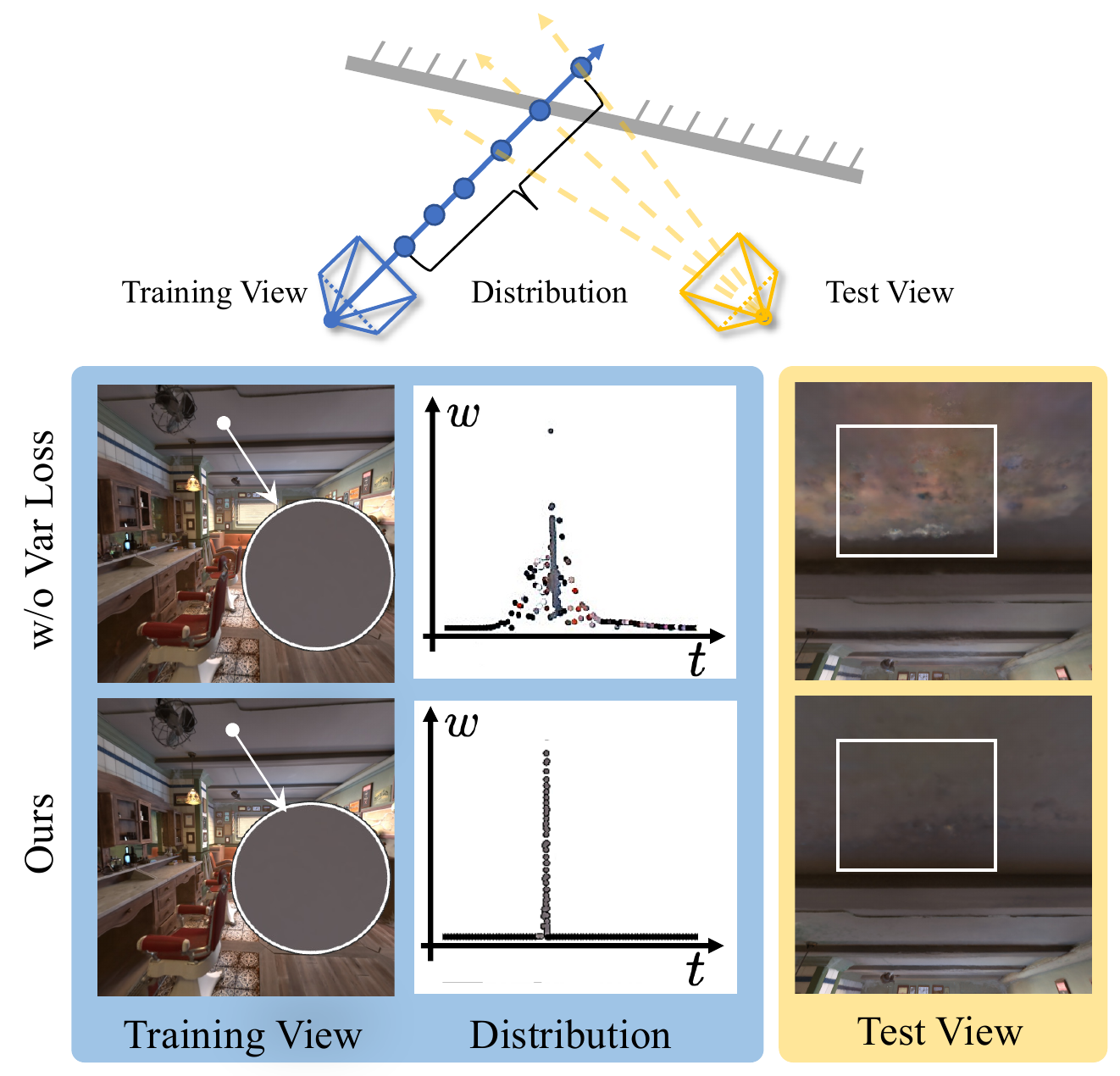}
\caption{\textbf{Variance regularization.} 
We show that even with the same expected depth and color, the density and color distribution can vary significantly.
This arbitrary distribution may not incur errors among interpolation tasks, while causing notable artifacts in extrapolation. Our NeRFVS uses variance regularization to reduce this ambiguity and produces significantly better visual quality.
}
\vspace{-\baselineskip}
\label{fig:var_demo}
\end{figure}

To decrease the ambiguity of low-texture and few-shot regions, we introduce the regularization on the variance of weight distribution $w(\mathbf{r})$ and color distribution $\mathbf{c}(\mathbf{r}, \mathbf{d})$. 
$\hat{D}(\mathbf{r})$ from volume rendering is a weighted sum of sample distance $t$.
Thus, without sufficient constraint signal, even if the $\hat{D}(\mathbf{r})$ is identical, the weight distribution can vary significantly, as shown in Fig.~\ref{fig:var_demo}.
Unfortunately, low texture and few-shot regions are common in indoor scene data.
Thus, we propose to further regularize the variance of density and color distribution to inhibit the ambiguity of these regions. 
Our variance regularization on weights is defined as follows:
\begin{align}\label{eq:varw loss}
    \mathcal{L}^{\text{w}}_{\text{var}}(\mathbf{r}) = \int_{t_{n}}^{t_{f}} w(\mathbf{r}(t))\left(t-\hat{D}(r)\right)^{2} dt
\end{align}
With the same weight distribution $w(\mathbf{r})$ and rendered pixel color $\hat{\mathbf{C}}(\mathbf{r})$, the color of each discrete point can still vary significantly.
This issue has little influence on the interpolation performance since the rays emitted at similar angles of the training set result in similar RGB values. However, when the novel view is significantly different from the training views, the derived color value $\mathbf{c}(\mathbf{r}(t),\mathbf{d})$ is messy and will cause unexpected artifacts, as shown in Fig.~\ref{fig:var_demo}.
To address this issue, we design a variance loss on color termed $\mathcal{L}^{\text{c}}_{\text{var}}(\mathbf{r})$, which is defined as follows:
\begin{align}\label{eq:varc loss}
    \mathcal{L}^{\text{c}}_{\text{var}}(\mathbf{r}) = \int_{t_{n}}^{t_{f}} \mathcal{K}(w(\mathbf{r}(t)))\left(\mathbf{c}(\mathbf{r}(t),\mathbf{d})-\hat{\mathbf{C}}(\mathbf{r})\right)^{2} dt
\end{align}
where $\mathcal{K}(\cdot)$ means stopping gradient.


\begin{figure*}[ht]
\centering
    \includegraphics[width=0.94 \textwidth,trim={0cm 0cm ,0cm 0cm}]{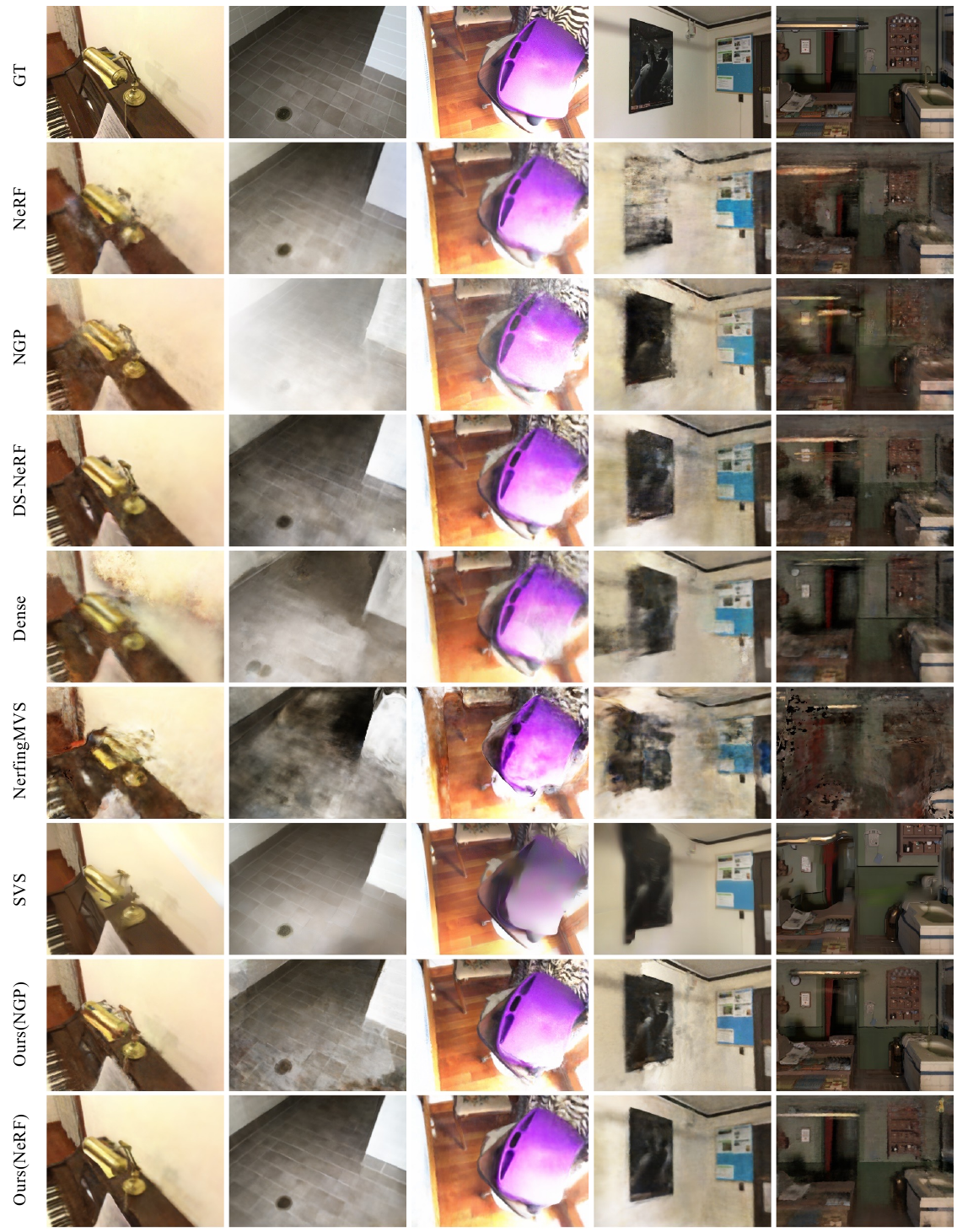}
    \caption{\textbf{Qualitative results.} 
    From top to bottom, we show the ground truth and extrapolation rendering results obtained by different methods on the ScanNet and Barbershop datasets.
    }
    \label{fig:qualitative_results}
\end{figure*}



\subsection{View Coverage Adjustment}
\label{ssec: View Count Adjustment}
Performing indoor scene FVS encounters various regions with different attributes including texture, transparency, reflection and view coverage.
For instance, when modeling the ceiling which is low texture, Lambertian surface and with few observations, we ought to rely more on the pseudo depth from the geometry scaffold and increase the variance constraint to reduce the ambiguity under photometric loss.


To address the view imbalance, we design an optimizing strategy based on the view coverage to adjust the intensity of the depth and variance loss. Specifically, we first use ray tracing and shadow mapping to obtain the view coverage map of each image from the geometry scaffold.
Then we apply different regularization weights accordingly based on the view coverage.
The optimizing strategy is defined as follows:
\begin{align}\label{eq:view coverage adjustment}
\lambda(\mathbf{r})=\left\{\begin{array}{ll}
1, & \text {if} \quad V(\mathbf{r})>\alpha \\
1+\frac{\lambda_{max}-1}{\alpha-1}(\alpha-V(\mathbf{r})),&\text{otherwise}
\end{array}\right.
\end{align}
where $V(\mathbf{r})$ is the view coverage of the ray $\mathbf{r}$,  $\lambda(\mathbf{r})$ represents the loss term of the depth and variance regularization, $\alpha$ and $\lambda_{max}$ are constants for view coverage adjustment. 
We strongly regularize the depth and variance on
areas whose number of observations is less than $\alpha$, serving as a powerful supplement to the photometric loss (Eq.~\ref{eq:photometric_loss}).

In summary, the total loss for optimizing each ray is:
\begin{align}
\label{eq:total_loss}
\begin{split}
\mathcal{L}(\mathbf{r}) &= \mathcal{L}_{\text{color}}(\mathbf{r})\\& + \lambda(\mathbf{r})(\lambda_{\text{d}}\mathcal{L}_{\text{depth}}(\mathbf{r})+\lambda_{\text{w}}\mathcal{L}_{\text{var}}^{\text{w}}(\mathbf{r})+\lambda_{\text{c}}\mathcal{L}_{\text{var}}^{\text{c}}(\mathbf{r}))
\end{split}
\end{align}
which is a linear combination of all losses presented above with loss weights $\lambda_{\text{d}}$, $\lambda_{\text{w}}$ and $\lambda_{\text{c}}$.


\begin{table*}[t]
    \centering
    \resizebox{1\linewidth}{!}{
        {\begin{tabular}{rrrrrrrrrrrrr}
\toprule
\multicolumn{1}{c}{} & \multicolumn{3}{c}{Scene0050\_00} & \multicolumn{3}{c}{Scene0084\_00} & \multicolumn{3}{c}{Scene0580\_00} & \multicolumn{3}{c}{Scene0616\_00} \\
\cmidrule(lr){2-4}
\cmidrule(lr){5-7}
\cmidrule(lr){8-10}
\cmidrule(lr){11-13}
 \multicolumn{1}{l}{\textbf{Extrapolation setting}} & $\uparrow$PSNR & $\uparrow$SSIM & $\downarrow$LPIPS & $\uparrow$PSNR & $\uparrow$SSIM & $\downarrow$LPIPS& $\uparrow$PSNR & $\uparrow$SSIM & $\downarrow$LPIPS & $\uparrow$PSNR & $\uparrow$SSIM & $\downarrow$LPIPS \\
\midrule
\multicolumn{1}{l}{{\color[rgb]{0,0,0} NeRF \cite{nerf}}} & 21.625 & 0.741 & 0.450 & 20.017 & 0.870 & 0.426     & 22.421 & 0.773 & 0.466 & 18.967 & 0.781 & 0.481  \\ 
\multicolumn{1}{l}{{\color[rgb]{0,0,0} Instant-NGP \cite{ngp}}} & 20.930 & 0.584 & 0.425 & 15.176 & 0.691 & 0.534     & 19.525 & 0.579 & 0.480 & 18.073 & 0.620 & 0.487  \\

\multicolumn{1}{l}{{\color[rgb]{0,0,0} DS-NeRF \cite{dsnerf}}} & 22.848 & 0.764 & 0.429 & 20.220 & 0.858 & 0.455     & 22.665 & 0.778 & 0.466 & 19.796 & 0.794 & 0.483  \\ 
\multicolumn{1}{l}{{\color[rgb]{0,0,0} Dense Depth Priors \cite{dense}}} & 22.800 & 0.614 & 0.426 & 19.232 & 0.781 & 0.418     & 22.399 & 0.634 & 0.460 & 19.986 & 0.684 & 0.442  \\ 
\multicolumn{1}{l}{{\color[rgb]{0,0,0} NerfingMVS \cite{nerfingmvs}}} & 16.893 & 0.613 & 0.600 & 13.975 & 0.699 & 0.645     & 18.106 & 0.662 & 0.606 & 14.569 & 0.645 & 0.663  \\ 
\multicolumn{1}{l}{{\color[rgb]{0,0,0} SVS \cite{svs}}} & 21.231 & 0.732 & {\bf 0.314} & 16.878 & 0.859 & {\bf 0.367}     & 21.575 & 0.783 & {\bf 0.319} & 17.530 & 0.795 & {\bf 0.406}  \\

\multicolumn{1}{l}{{\color[rgb]{0,0,0} Ours(NGP)}} & 22.479 & 0.623 & 0.357 & 20.553 & 0.723 & 0.451     & 22.043 & 0.634 & 0.390 & 19.352 & 0.636 & 0.473  \\
\multicolumn{1}{l}{{\color[rgb]{0,0,0} Ours(NeRF)}} & {\bf 24.261} & {\bf 0.788} & 	0.388 & {\bf 24.452} & {\bf 0.899} & 0.401     & {\bf 23.964} & {\bf 0.798} & 0.434 & {\bf 22.180} & {\bf 0.832} & 0.433  \\
\midrule

 \multicolumn{1}{l}{\textbf{Interpolation setting}} & $\uparrow$PSNR & $\uparrow$SSIM & $\downarrow$LPIPS & $\uparrow$PSNR & $\uparrow$SSIM & $\downarrow$LPIPS& $\uparrow$PSNR & $\uparrow$SSIM & $\downarrow$LPIPS & $\uparrow$PSNR & $\uparrow$SSIM & $\downarrow$LPIPS \\
\midrule
\multicolumn{1}{l}{{\color[rgb]{0,0,0} NeRF \cite{nerf}}} & 25.215          & 0.818          & 0.370          & 26.953          & 0.927          & 0.356           & 25.672          & 0.828          & 0.430          & 23.278          & 0.840          & 0.407           \\ 

\multicolumn{1}{l}{{\color[rgb]{0,0,0} Instant-NGP \cite{ngp}}} & 24.505          & 0.675          & 0.318          & 21.742          & 0.800          & 0.378           & 23.657          & 0.677          & 0.379          & 21.718          & 0.681          & 0.391           \\
\multicolumn{1}{l}{{\color[rgb]{0,0,0} DS-NeRF \cite{dsnerf}}} & 25.678 & 0.822          & 0.368          & 27.879          & 0.927          & 0.365           & 26.129 & 0.832          & 0.431          & 24.124          & 0.839          & 0.428           \\
\multicolumn{1}{l}{{\color[rgb]{0,0,0} Dense Depth Priors \cite{dense}}} & 24.971          & 0.668          & 0.371          & 22.580          & 0.820          & 0.384           & 24.995          & 0.682          & 0.432          & 21.322          & 0.697          & 0.415           \\ 
\multicolumn{1}{l}{{\color[rgb]{0,0,0} NerfingMVS \cite{nerfingmvs}}} & 22.020          & 0.751          & 0.459          & 22.789          & 0.856          & 0.486           & 22.384          & 0.766          & 0.511          & 19.824          & 0.754          & 0.530           \\
\multicolumn{1}{l}{{\color[rgb]{0,0,0} SVS \cite{svs}}} & 23.937          & \textbf{0.830} & \textbf{0.239} & 18.899          & 0.901          & \textbf{0.302} & 24.308          & \textbf{0.844} & \textbf{0.252} & 19.443          & 0.829          & \textbf{0.341}  \\

\multicolumn{1}{l}{{\color[rgb]{0,0,0} Ours(NGP)}} & 25.319          & 0.699          & 0.286          & 25.705          & 0.820          & 0.347           & 25.303          & 0.709          & 0.336          & 22.960          & 0.702          & 0.374           \\
\multicolumn{1}{l}{{\color[rgb]{0,0,0} Ours(NeRF)}} & \textbf{25.840}          & 0.828          & 0.347          & \textbf{28.004} & \textbf{0.929} & 0.360           & \textbf{26.157}          & 0.835          & 0.412          & \textbf{24.362} & \textbf{0.850} & 0.407           \\

\bottomrule

\end{tabular}}} 
    \vspace{-0.5em}
    \caption{
        \textbf{Quantitative results on ScanNet dataset.} We compare our method with NeRF-like and
other rendering methods. Our method can significantly improve the performance of both \nerf and instant-NGP.
        }
    \label{tab:quantitative_results}
    \vspace{-\baselineskip}
\end{table*}

\begin{table}[t]
\centering
\resizebox{\linewidth}{!}{
{\begin{tabular}{lcccccc}
\toprule
\multicolumn{1}{c}{} & \multicolumn{3}{c}{Extrapolation} & \multicolumn{3}{c}{Interpolation} \\
\cmidrule(lr){2-4}
\cmidrule(lr){5-7}
\multicolumn{1}{l}{Barbershop} & $\uparrow$PSNR & $\uparrow$SSIM & $\downarrow$LPIPS &$\uparrow$PSNR & $\uparrow$SSIM & $\downarrow$LPIPS  \\
\midrule
NeRF\cite{nerf}     & 24.367 & 0.868 & 0.294  & 32.416 & 0.957 & 0.149 \\ 
Instant-NGP\cite{ngp} & 19.712  & 0.753  & 0.397   & \textbf{36.121}  & 0.964  & 0.056   \\
DS-NeRF\cite{dsnerf}  & 24.105  & 0.866  & 0.313 & 32.081  & 0.952  & 0.174    \\ 
Dense Depth Priors \cite{dense} & 20.700  & 0.723  & 0.470  & 29.074  & 0.837  & 0.284  \\ 
NerfingMVS \cite{nerfingmvs} & 15.577  & 0.616  & 0.635 & 27.265  & 0.897  & 0.247  \\ 
SVS\cite{svs} & 21.725 & 0.878 & \textbf{0.210} & 32.997 & \textbf{0.988} & \textbf{0.019}  \\ 
Ours(NGP)       & 24.276 & 0.833 & 0.235 & 33.756 & 0.959 & 0.052 \\
Ours(NeRF)       & \textbf{26.946} & \textbf{0.891} & 0.268  & 32.766 & 0.958 & 0.146\\

\bottomrule
\end{tabular} 				

}}

\caption{\textbf{Quantitative results on Barbershop dataset.}
}
\label{tab:barbershop_eval}
\vspace{-\baselineskip}
\end{table}




\cvprsection{Experiments}
We conduct a series of experiments to evaluate our method and to test whether the proposed modules enable better FVS performance among indoor scenes.


\subsection{Datasets}
\noindent \textbf{Barbershop.}
Our FVS task aims to enable 6-DOF navigation in indoor scenes, but evaluating FVS requires considering all candidate view directions and positions, which is difficult to achieve with real-world scenes. To address this, we created the Barbershop~\cite{barbershop} FVS dataset using Blender's Cycles pathtracer~\cite{blender}, including interpolation and extrapolation views to enable evaluation. Barbershop consists of 543 images captured in a human-like trajectory. For interpolation, we used one frame from every six frames. For extrapolation, we grid-sampled camera positions and assigned 24 evenly distributed directions to each position.

\noindent \textbf{ScanNet.}
ScanNet\cite{scannet} is a large-scale indoor dataset with 1613 scenes, including ground truth camera intrinsics and camera poses. 
We choose the scenes chosen in~\cite{manhattansdf} for experiments. In particular, we select the middle frames between training frames as nvs frames, which can be seen as interpolations of the training set. For our extrapolation setting, we select several trajectories different from that in the training set to build up a more difficult test set. The ratio of the training, interpolation and extrapolation is 2:1:1. We show a real trajectory from ``scene0050\_00'' in Fig~.\ref{fig:view coverage} (b) for further illustration.

\subsection{Implementation Details and Metrics}
Our implementation is based on MindSpore~\cite{mindspore}, which is a new deep-learning computing framework. The training is performed on one NVIDIA RTX3090 GPU. In experiments, we normalize all cameras to be inside a unit sphere so that the scene is bounded within $\left[-1, 1\right]^3$. The image resolution in ScanNet is 468×624 after resizing and cropping dark borders, following~\cite{dense}. We train our model using Adam optimizer with an initial learning rate of 1e-3 and train the network with batches of 1024 rays for 20k iterations. 
$\alpha$ is set to 9, and $\lambda_{max}$ is set to 5 throughout all the experiments except for relative ablation.
Besides, we prepare a relaxing stage which occupies 10\% of training iterations to fine-tune the model.
For further implementation details, please refer to the supplementary. 

We use visual quality assessment metrics to evaluate the performance of Free View Synthesis, including the PSNR, the SSIM\cite{ssim}, and the LPIPS\cite{lpips}.


\subsection{Comparisons with the State-of-the-art Methods}
We construct our NeRFVS based on vanilla NeRF~\cite{nerf}, term as Ours(NeRF) and instant-NGP~\cite{ngp}, term as Ours(NGP) and make comparisons with NeRF-like and other rendering methods, including a) Neural volume rendering methods:  NeRF and instant-NGP; b) a method with sparse depth input: DS-NeRF\cite{dsnerf} which uses colmap~\cite{colmap} to acquire sparse depth priors; c) methods with depth completion: Dense Depth Priors\cite{dense} and NerfingMVS\cite{nerfingmvs}, which get dense depth priors from sparse point clouds with completion networks; d) a method with on-surface image feature aggregation: Stable View Synthesis (SVS)\cite{svs}, which maps view-dependent features extracted from input images to surfaces and generates images from surface features. 

In Table~\ref{tab:quantitative_results} and Table~\ref{tab:barbershop_eval}, we report the interpolation and extrapolation results across the test images in two datasets.
Qualitative results are shown in Figure~\ref{fig:qualitative_results}.
Comparing the baselines and ours, our method significantly improves the extrapolation ability of both NeRF and instant-NGP. Ours(NeRF) can synthesize high-fidelity images with a consistent 3D layout compared to the baseline. Integrating holistic priors and regularization on radiance distribution into the learning of a neural radiance field efficiently reduces the floaters and distortions, which is a common problem in applying NeRF with indoor scenes, resulting in clean and reasonable FVS performances. Considering Ours(NeRF) produces competitive performances among interpolation settings and significantly outperforms other methods among extrapolation settings considering both PSNR and SSIM, our method achieves new state-of-art performance among indoor scene FVS tasks. 
The extrapolation ability of instant-NGP is weaker than NeRF, as shown in Table~\ref{tab:quantitative_results}, while Ours(NGP) still achieves competitive performance. Considering Ours(NGP) trains and inferences significantly faster than NeRF, about $8\times$ speedup, a spend-quality trade-off exists.

DS-NeRF leverages sparse depth with re-projection error as confidence, which is error-prone compared with the holistic priors since the sparse depth may omit components that rarely appear in training views (example 4, 5, Fig.~\ref{fig:qualitative_results}). 
NerfingMVS and Dense leverage depth completion network to generate per-frame depth maps, which are not 3D consistent compared with our geometry scaffold. Besides, the view imbalance property makes the depth-completion performance vary dramatically, leading to unstable optimization (example 1, 2, Fig.~\ref{fig:qualitative_results}). 
Our method takes the view imbalance and ray regularization into consideration, resulting in better performance on few-shot regions. 

SVS performs well on LPIPS while synthesizing noticeable artifacts and distortions (example 3, 4, 5, Fig.~\ref{fig:qualitative_results}), which significantly degrades the experience of indoor roaming.
The superior LPIPS performance may be attributed to the direct supervision of an LPIPS-like perceptual loss, as discussed in~\cite{barron2022mip}, while our method only minimizes per-pixel loss.
SVS heavily relies on the quality of the geometry scaffold.
When the geometry scaffold is inaccurate, the renderings will be distorted.
Besides, SVS runs COLMAP MVS on training and testing images together to get their geometry scaffold, which gives them an advantage. Our method only relies on training images and applies a robust depth loss to alleviate the influence of the inaccurate geometry, leading to more consistent and high-fidelity results.



\subsection{Analysis}
We show that each proposed module is of critical importance to the final rendering performance in Table~\ref{tab:ablation}. We also discuss the influence of different geometry scaffold sources.

\noindent \textbf{Without Depth Loss.}
In this experiment, we replace our robust depth loss in Eq. (\ref{eq:rubust depth loss}) with L2 loss. 
As aforementioned in Sec.~\ref{ssec: Robust Depth Loss}, our robust depth loss can alleviate the influence of pseudo depth error. L2 loss struggles to deal with the inaccuracy of geometry, causing artifacts and disappearance of objects, \textit{e.g.}, lamps, as shown in Fig.~\ref{fig:robust_loss}.


\noindent \textbf{Without Var Loss.}
We omit the variance loss, including the $\mathcal{L}^{\text{w}}_{\text{var}}(\mathbf{r})$ and $\mathcal{L}^{\text{c}}_{\text{var}}(\mathbf{r})$  (Eq.~(\ref{eq:varw loss}), Eq. (\ref{eq:varc loss})).
Even with consistent and dense depth priors, only depending on the depth expectation to guide the optimization of a NeRF over poorly textured regions or few-shot regions (\textit{e.g.} ceiling) is particularly challenging since the density and color distribution is arbitrary.
This results in various artifacts when performing extrapolation, as shown in Fig~\ref{fig:var_demo}.

\noindent \textbf{Without Adjustment.}
We remove the view coverage adjustment in this experiment, which means we treat all areas as fully observed and apply weak depth and variance regularization. 
In Table~\ref{tab:ablation}, the quantitative results show the degraded performance because of unbalanced regularization.

\begin{figure}[t]
  \centering
  \begin{subfigure}[b]{0.49\linewidth}
    \includegraphics[width=\linewidth]{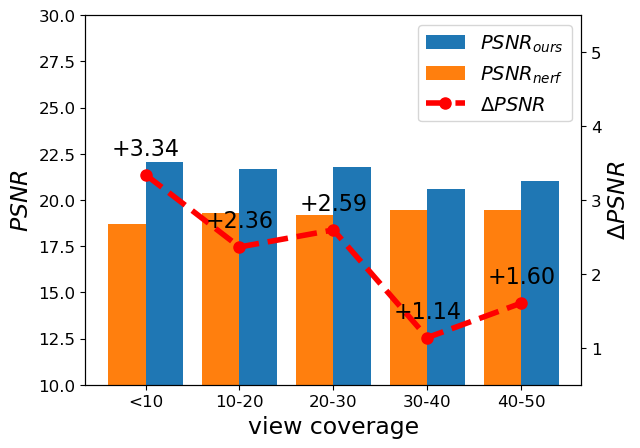}
  \end{subfigure}
  \hfill %
  \begin{subfigure}[b]{0.49\linewidth}
    \includegraphics[width=\linewidth]{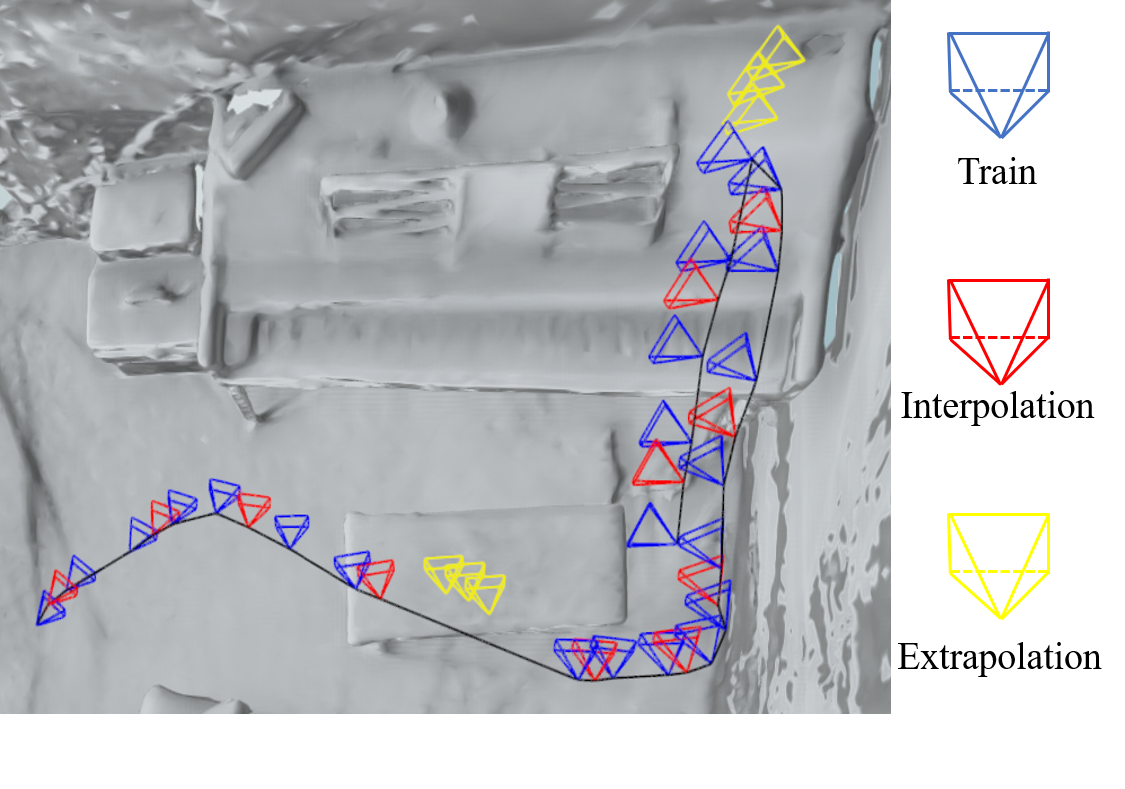}
  \end{subfigure}
  \caption{Left(a): Performance along the view coverage. Right(b): Trajectory with train, interpolation, and extrapolation viewpoints.}
  \label{fig:view coverage}
\end{figure}

\begin{table}[t]
\centering
\resizebox{0.8 \linewidth}{!}{
{






\begin{tabular}{lccc}
\toprule
& $\uparrow$PSNR & $\uparrow$SSIM & $\downarrow$LPIPS  \\
\midrule
ManhattanSDF~\cite{manhattansdf}    & 19.776 & 0.775 & 0.570  \\
NeRF~\cite{nerf}               & 20.758 & 0.791 & 0.456  \\
w/o Depth Loss     & 23.187&0.821&0.428  \\
w/o Var Loss   & 23.435&0.822&0.431 \\
w/o Adjustment & 23.172&0.821&0.429  \\
\midrule
DS-NeRF~\cite{dsnerf} + Colmap MVS & 20.906 & 0.790 & 0.485  \\
DS-NeRF + Our depth   & 21.385 & 0.800 & 0.464 \\
Ours + Colmap MVS       & 22.771 & 0.817 & 0.439  \\
Full model       & 23.714 & 0.829 & 0.414  \\

\bottomrule
\end{tabular}}} 
\caption{\textbf{Ablation studies on ScanNet.} We report the mean image quality metrics among extrapolation setting with Ours(NeRF).
}
\label{tab:ablation}
\vspace{-\baselineskip}
\end{table}
\noindent \textbf{Performance Gain with View Coverage.} We evaluated the effectiveness of NeRFVS in terms of view coverage, as shown in Fig.~\ref{fig:view coverage}(a). The results demonstrate improvement over all extrapolation results, particularly in regions with insufficient view coverage, compared to NeRF.

\noindent \textbf{Improvement over the scaffold.} 
Experiments in Table. \ref{tab:ablation} shows that NeRFVS significantly outperforms DS-NeRF with our depth. Besides, DS-NeRF gains minor improvement by replacing the depth from Colmap MVS with our depth.
These results prove that the improvement over prior work mainly depends on the method, rather than the choice of the scaffold.
Though Colmap MVS is not suitable for our NeRFVS, ours with Colmap MVS still outperforms DS-NeRF with Colmap MVS, showing that our method is generic and can be applied to different scaffolds.

\cvprsection{Conclusion}
In this study, we introduce NeRFVS, a novel method enabling NeRF to perform free navigation for indoor scenes.
Our method utilizes the holistic prior from a neural-reconstructed geometric scaffold to guide the optimization of NeRF and boosts the rendering accuracy by introducing the depth and variance constraints with dynamical adjustment.
We show that the inaccuracy of the geometry scaffold degrades the extrapolation performance and propose a robust depth loss to address this issue. 
Variance regularization and view coverage-based training strategy improve the rendering quality on the few shot regions.
Experiments show that NeRFVS significantly improves the extrapolation ability of NeRF-based methods. We believe our method leaps a meaningful step toward extending NeRF to free navigation.

\noindent \textbf{Limitations.}
NeRFVS reduces geometry errors in regions with sufficient observations but relies on pseudo depth in few-shot regions. This may result in poor extrapolation if the geometry is exceptionally poor.
The way we generate the view coverage map is an adequate approximation in plainer regions. While in some extreme cases, such as holes and thin objects, this could result in errors. Besides, our view coverage only considers the number of observations, regardless of the emitting directions, which may cause degradation in view-dependent appearance.

\noindent\textbf{Acknowledgments.} This work was supported by NSFC 62176159, Natural Science Foundation of Shanghai 21ZR1432200, Shanghai Municipal Science and Technology Major Project 2021SHZDZX0102, and the Fundamental Research Funds for the Central Universities. We thank Ruofan Liang for inspiring and valuable discussions. We thank MindSpore~\cite{mindspore} for the partial support to this work, which is a new deep learning computing framework.

{\small
\bibliographystyle{ieee_fullname}
\bibliography{egbib}

\begin{thebibliography}{10}\itemsep=-1pt

\bibitem{barbershop}
Barbershop.
\newblock
  \url{https://svn.blender.org/svnroot/bf-blender/trunk/lib/benchmarks/cycles/barbershop_interior/}.

\bibitem{barron2022mip}
Jonathan~T Barron, Ben Mildenhall, Dor Verbin, Pratul~P Srinivasan, and Peter
  Hedman.
\newblock Mip-nerf 360: Unbounded anti-aliased neural radiance fields.
\newblock In {\em Proceedings of the IEEE/CVF Conference on Computer Vision and
  Pattern Recognition}, pages 5470--5479, 2022.

\bibitem{ulr}
Chris Buehler, Michael Bosse, Leonard McMillan, Steven Gortler, and Michael
  Cohen.
\newblock Unstructured lumigraph rendering.
\newblock In {\em Proceedings of the 28th Annual Conference on Computer
  Graphics and Interactive Techniques}, SIGGRAPH '01, page 425–432, New York,
  NY, USA, 2001. Association for Computing Machinery.

\bibitem{blender}
Blender~Online Community.
\newblock Blender - a 3d modelling and rendering package, 2018.

\bibitem{scannet}
Angela Dai, Angel~X Chang, Manolis Savva, Maciej Halber, Thomas Funkhouser, and
  Matthias Nie{\ss}ner.
\newblock Scannet: Richly-annotated 3d reconstructions of indoor scenes.
\newblock In {\em Proceedings of the IEEE conference on computer vision and
  pattern recognition}, pages 5828--5839, 2017.

\bibitem{deng2022depth}
Kangle Deng, Andrew Liu, Jun-Yan Zhu, and Deva Ramanan.
\newblock Depth-supervised nerf: Fewer views and faster training for free.
\newblock In {\em Proceedings of the IEEE/CVF Conference on Computer Vision and
  Pattern Recognition}, pages 12882--12891, 2022.

\bibitem{dsnerf}
Kangle Deng, Andrew Liu, Jun-Yan Zhu, and Deva Ramanan.
\newblock Depth-supervised nerf: Fewer views and faster training for free.
\newblock In {\em Proceedings of the IEEE/CVF Conference on Computer Vision and
  Pattern Recognition}, pages 12882--12891, 2022.

\bibitem{fastrcnn}
Ross Girshick.
\newblock Fast r-cnn.
\newblock In {\em Proceedings of the IEEE international conference on computer
  vision}, pages 1440--1448, 2015.

\bibitem{manhattansdf}
Haoyu Guo, Sida Peng, Haotong Lin, Qianqian Wang, Guofeng Zhang, Hujun Bao, and
  Xiaowei Zhou.
\newblock Neural 3d scene reconstruction with the manhattan-world assumption.
\newblock In {\em Proceedings of the IEEE/CVF Conference on Computer Vision and
  Pattern Recognition}, pages 5511--5520, 2022.

\bibitem{huberloss}
Peter~J Huber.
\newblock Robust estimation of a location parameter.
\newblock In {\em Breakthroughs in statistics}, pages 492--518. Springer, 1992.

\bibitem{kim2022infonerf}
Mijeong Kim, Seonguk Seo, and Bohyung Han.
\newblock Infonerf: Ray entropy minimization for few-shot neural volume
  rendering.
\newblock In {\em Proceedings of the IEEE/CVF Conference on Computer Vision and
  Pattern Recognition}, pages 12912--12921, 2022.

\bibitem{nerf}
Ben Mildenhall, Pratul~P Srinivasan, Matthew Tancik, Jonathan~T Barron, Ravi
  Ramamoorthi, and Ren Ng.
\newblock Nerf: Representing scenes as neural radiance fields for view
  synthesis.
\newblock {\em Communications of the ACM}, 65(1):99--106, 2021.

\bibitem{ngp}
Thomas M{\"u}ller, Alex Evans, Christoph Schied, and Alexander Keller.
\newblock Instant neural graphics primitives with a multiresolution hash
  encoding.
\newblock {\em arXiv preprint arXiv:2201.05989}, 2022.

\bibitem{niemeyer2022regnerf}
Michael Niemeyer, Jonathan~T Barron, Ben Mildenhall, Mehdi~SM Sajjadi, Andreas
  Geiger, and Noha Radwan.
\newblock Regnerf: Regularizing neural radiance fields for view synthesis from
  sparse inputs.
\newblock In {\em Proceedings of the IEEE/CVF Conference on Computer Vision and
  Pattern Recognition}, pages 5480--5490, 2022.

\bibitem{philip2021free}
Julien Philip, S{\'e}bastien Morgenthaler, Micha{\"e}l Gharbi, and George
  Drettakis.
\newblock Free-viewpoint indoor neural relighting from multi-view stereo.
\newblock {\em ACM Transactions on Graphics (TOG)}, 40(5):1--18, 2021.

\bibitem{fvs}
Gernot Riegler and Vladlen Koltun.
\newblock Free view synthesis.
\newblock In {\em European Conference on Computer Vision}, pages 623--640.
  Springer, 2020.

\bibitem{svs}
Gernot Riegler and Vladlen Koltun.
\newblock Stable view synthesis.
\newblock In {\em Proceedings of the IEEE/CVF Conference on Computer Vision and
  Pattern Recognition}, pages 12216--12225, 2021.

\bibitem{roessle2022dense}
Barbara Roessle, Jonathan~T Barron, Ben Mildenhall, Pratul~P Srinivasan, and
  Matthias Nie{\ss}ner.
\newblock Dense depth priors for neural radiance fields from sparse input
  views.
\newblock In {\em Proceedings of the IEEE/CVF Conference on Computer Vision and
  Pattern Recognition}, pages 12892--12901, 2022.

\bibitem{dense}
Barbara Roessle, Jonathan~T Barron, Ben Mildenhall, Pratul~P Srinivasan, and
  Matthias Nie{\ss}ner.
\newblock Dense depth priors for neural radiance fields from sparse input
  views.
\newblock In {\em Proceedings of the IEEE/CVF Conference on Computer Vision and
  Pattern Recognition}, pages 12892--12901, 2022.

\bibitem{colmap}
Johannes~L Schonberger and Jan-Michael Frahm.
\newblock Structure-from-motion revisited.
\newblock In {\em Proceedings of the IEEE conference on computer vision and
  pattern recognition}, pages 4104--4113, 2016.

\bibitem{imap}
Edgar Sucar, Shikun Liu, Joseph Ortiz, and Andrew~J Davison.
\newblock imap: Implicit mapping and positioning in real-time.
\newblock In {\em Proceedings of the IEEE/CVF International Conference on
  Computer Vision}, pages 6229--6238, 2021.

\bibitem{neuralrecon}
Jiaming Sun, Yiming Xie, Linghao Chen, Xiaowei Zhou, and Hujun Bao.
\newblock Neuralrecon: Real-time coherent 3d reconstruction from monocular
  video.
\newblock In {\em Proceedings of the IEEE/CVF Conference on Computer Vision and
  Pattern Recognition}, pages 15598--15607, 2021.

\bibitem{sun2022learning}
Yuqi Sun, Shili Zhou, Ri Cheng, Weimin Tan, Bo Yan, and Lang Fu.
\newblock Learning robust image-based rendering on sparse scene geometry via
  depth completion.
\newblock In {\em Proceedings of the IEEE/CVF Conference on Computer Vision and
  Pattern Recognition}, pages 7813--7823, 2022.

\bibitem{mindspore}
Huawei Technologies.
\newblock Mindspore.
\newblock \url{https://www.mindspore.cn/}.

\bibitem{neuris}
Jiepeng Wang, Peng Wang, Xiaoxiao Long, Christian Theobalt, Taku Komura,
  Lingjie Liu, and Wenping Wang.
\newblock Neuris: Neural reconstruction of indoor scenes using normal priors.
\newblock In {\em Computer Vision--ECCV 2022: 17th European Conference, Tel
  Aviv, Israel, October 23--27, 2022, Proceedings, Part XXXII}, pages 139--155.
  Springer, 2022.

\bibitem{wang2021ibrnet}
Qianqian Wang, Zhicheng Wang, Kyle Genova, Pratul~P Srinivasan, Howard Zhou,
  Jonathan~T Barron, Ricardo Martin-Brualla, Noah Snavely, and Thomas
  Funkhouser.
\newblock Ibrnet: Learning multi-view image-based rendering.
\newblock In {\em Proceedings of the IEEE/CVF Conference on Computer Vision and
  Pattern Recognition}, pages 4690--4699, 2021.

\bibitem{ssim}
Zhou Wang, Alan~C Bovik, Hamid~R Sheikh, and Eero~P Simoncelli.
\newblock Image quality assessment: from error visibility to structural
  similarity.
\newblock {\em IEEE transactions on image processing}, 13(4):600--612, 2004.

\bibitem{nerfingmvs}
Yi Wei, Shaohui Liu, Yongming Rao, Wang Zhao, Jiwen Lu, and Jie Zhou.
\newblock Nerfingmvs: Guided optimization of neural radiance fields for indoor
  multi-view stereo.
\newblock In {\em Proceedings of the IEEE/CVF International Conference on
  Computer Vision}, pages 5610--5619, 2021.

\bibitem{wu2022scalable}
Xiuchao Wu, Jiamin Xu, Zihan Zhu, Hujun Bao, Qixing Huang, James Tompkin, and
  Weiwei Xu.
\newblock Scalable neural indoor scene rendering.
\newblock {\em ACM Transactions on Graphics (TOG)}, 41(4):1--16, 2022.

\bibitem{xu2022sinnerf}
Dejia Xu, Yifan Jiang, Peihao Wang, Zhiwen Fan, Humphrey Shi, and Zhangyang
  Wang.
\newblock Sinnerf: Training neural radiance fields on complex scenes from a
  single image.
\newblock {\em arXiv preprint arXiv:2204.00928}, 2022.

\bibitem{xu2021scalable}
Jiamin Xu, Xiuchao Wu, Zihan Zhu, Qixing Huang, Yin Yang, Hujun Bao, and Weiwei
  Xu.
\newblock Scalable image-based indoor scene rendering with reflections.
\newblock {\em ACM Transactions on Graphics (TOG)}, 40(4):1--14, 2021.

\bibitem{yeh2022photoscene}
Yu-Ying Yeh, Zhengqin Li, Yannick Hold-Geoffroy, Rui Zhu, Zexiang Xu,
  Milo{\v{s}} Ha{\v{s}}an, Kalyan Sunkavalli, and Manmohan Chandraker.
\newblock Photoscene: Photorealistic material and lighting transfer for indoor
  scenes.
\newblock In {\em Proceedings of the IEEE/CVF Conference on Computer Vision and
  Pattern Recognition}, pages 18562--18571, 2022.

\bibitem{monosdf}
Zehao Yu, Songyou Peng, Michael Niemeyer, Torsten Sattler, and Andreas Geiger.
\newblock Monosdf: Exploring monocular geometric cues for neural implicit
  surface reconstruction.
\newblock {\em ArXiv}, abs/2206.00665, 2022.

\bibitem{rapnerf}
Jian Zhang, Yuanqing Zhang, Huan Fu, Xiaowei Zhou, Bowen Cai, Jinchi Huang,
  Rongfei Jia, Binqiang Zhao, and Xing Tang.
\newblock Ray priors through reprojection: Improving neural radiance fields for
  novel view extrapolation.
\newblock In {\em Proceedings of the IEEE/CVF Conference on Computer Vision and
  Pattern Recognition}, pages 18376--18386, 2022.

\bibitem{lpips}
Richard Zhang, Phillip Isola, Alexei~A Efros, Eli Shechtman, and Oliver Wang.
\newblock The unreasonable effectiveness of deep features as a perceptual
  metric.
\newblock In {\em Proceedings of the IEEE conference on computer vision and
  pattern recognition}, pages 586--595, 2018.

\bibitem{Nice-slam}
Zihan Zhu, Songyou Peng, Viktor Larsson, Weiwei Xu, Hujun Bao, Zhaopeng Cui,
  Martin~R Oswald, and Marc Pollefeys.
\newblock Nice-slam: Neural implicit scalable encoding for slam.
\newblock In {\em Proceedings of the IEEE/CVF Conference on Computer Vision and
  Pattern Recognition}, pages 12786--12796, 2022.

\end{thebibliography}
}


\end{document}